# Acquisition d'informations lexicales à partir de corpus

Cédric Messiant, Thierry Poibeau
*Laboratoire d'Informatique de Paris-Nord*

**Introduction**
L'existence de gros corpus (plusieurs millions de mots) et d'analyseurs syntaxiques performants fait qu'il est actuellement possible d'extraire automatiquement des connaissances à large couverture sur les mots et les constructions associées, directement à partir de corpus. Cette démarche permet d'obtenir des lexiques très complets à moindre coût, avec également des informations sur la fréquence et la productivité de différentes constructions, c'est-à-dire des données difficilement calculables à la main.

Depuis une quinzaine d'années, plusieurs systèmes ont ainsi été conçus afin d'extraire automatiquement des informations sur la construction de mots essentiels du lexique, en général les verbes. On peut citer les travaux de (Brent (1993), Manning (1993), Briscoe and Carroll (1997), Korhonen (2002), Schulte im Walde (2002) parmi de nombreux autres. Nous avons nous-mêmes réalisé un système du même type pour le français, avec une première expérience qui s'appuie sur le corpus *Le Monde* (200 millions de mots, 1990–1999) et sur l'analyseur Syntex (Bourigault, 2007) pour inférer des connaissances sur la sous-catégorisation de plus de 3000 verbes (Messiant et Poibeau, 2008 ; Messiant 2008).

Le processus se décompose en 3 grandes étapes : 1) on rassemble d'abord l'ensemble des occurrences du verbe considéré ainsi que tous ses compléments, 2) on fait ensuite l'inventaire de toutes les constructions possibles pour le verbe considéré et enfin, 3) les constructions les plus rares sont éliminées, à partir de l'hypothèse qu'un nombre trop faible d'occurrences est le révélateur d'une erreur d'analyse (simple rencontre de surface). Tous les systèmes reposent sur cette architecture, même s'ils varient quant à la finesse de l'analyse considérée ou des stratégies de filtrage utilisées.

**Quelques difficultés récurrentes des approches à base de corpus**
Malgré les avantages décrits, l'approche n'est pas sans inconvénient. Comme elle se fonde sur des outils automatiques qui ne sont pas parfaits, les ressources ainsi constituées sont sujettes à erreur et doivent impérativement être révisées à la main. Il faut en outre un nombre d'occurrences suffisant pour qu'il soit possible d'inférer une information pertinente, ce qui veut dire qu'il y a un manque criant d'information pour tous les items lexicaux peu fréquents en corpus si on s'en tient à des analyses de ce type (cf. le "*sparse problem*" en anglais).

Par ailleurs, on voit que certaines constructions sont difficiles à capter tandis que d'autres brouillent l'analyse. D'un côté, les expressions semi-figées (cf. concept d'*entrenchment* dans les grammaires cognitives, Croft et Cruse 2004) ne sont pas repérées en tant que telles ; de l'autre, certains compléments sont fortement présents avec certains verbes : ils sont alors analysés comme arguments alors qu'il s'agit clairement de modifieurs. Enfin, les réalisations de surface ne permettent pas de différencier certains compléments (cf. …*donner un livre à Marie* vs …*donner un livre à Marseille*). Il y a là une divergence majeure avec la description linguistique qui met au premier plan ces différents éléments, particulièrement dans le cadre des linguistiques dites cognitives.

**Des pistes d'amélioration**
Nous pensons que ces problèmes ne remettent pas en cause l'acquisition à partir de corpus en tant que telle mais qu'ils montrent les limites des approches employées jusqu'à maintenant pour l'acquisition d'information sur la sous-catégorisation lexicale. Le fait de ne considérer que le verbe et des seuils sur les compléments rend l'approche rapidement opérationnelle et donne dans l'absolu une bonne approximation du problème considéré (à savoir, acquérir des informations sur la sous-catégorisation syntaxique). Cependant, le point de vue est très réducteur car il se limite à une vue partant du verbe, sans tenir compte d'autres contraintes extérieures.

Pour résoudre ces problèmes, il est donc nécessaire de complexifier le modèle de base. Nous considérons la langue comme étant soumises à un ensemble de contraintes qu'il est possible de modéliser statistiquement. Ce point n'est pas nouveau en soi et on le trouve sous de nombreuses formes dans la littérature (e.g. Shieber 1992 ; Blache 2001) . Cependant, sur le plan informatique, en acquisition de connaissances notamment, cette idée n'a été complètement explorée à ce jour.

Une meilleure prise en compte des contraintes à l'œuvre permettrait une modélisation plus fine de la sous-catégorisation verbale et ainsi une acquisition de meilleure qualité. Le lien entre le verbe et ses compléments, couramment pris en compte dans les approches ci-dessus comme nous l'avons montré, est pertinent mais insuffisant dans ce contexte. La prise en compte de la dispersion des têtes nominales des compléments ou des têtes des compléments prépositionnels permet de beaucoup mieux caractériser les séquences visées (c'est-à-dire qu'il est possible de calculer automatiquement la dispersion des noms en position de complément après un verbe donné, au à l'inverse un lien fort entre un verbe et un nom, pour former une expression plus ou moins figée comme *prendre en compte*— ceci permet de retrouver des colligations, pour reprendre le terme de Firth, 1968). En intégrant ces éléments lors de l'analyse, l'acquisition de schémas de sous-catégorisation devient beaucoup plus fine dans la mesure où le processus permet d'intégrer la notion de figement d'un côté, de liberté de déplacement et de caractérisation sémantique des modifieurs de l'autre (Fabre et Bourigault, 2008).

Enfin, l'approche ne peut se passer d'une validation manuelle. Plutôt que de reléguer celle-ci à un stade ultime, quand la machine ne peut plus rien, il semble beaucoup plus profitable de consulter l'utilisateur pendant le processus d'acquisition, afin de lui permettre de guider le processus lui-même. Il est en outre possible de repérer des verbes plus rares et de faire des propositions de cadres de sous-catégorisation, qui semblent pertinents à partir du moment où l'analyse repose sur quelques phrases dont l'analyse semble correcte (phrases courtes qui peuvent être analysées avec un bon degré de certitude). Enfin, en dernier recours, il est nécessaire de consulter directement l'utilisateur quand les connaissances disponibles en corpus restent insuffisantes.

**En guise de conclusion**

On ne saurait inférer un comportement cognitif à partir de ce schéma d'acquisition, mais on peut quand même remarquer la plausibilité d'un jeu de contraintes multiple, l'influence de la notion de fréquence, et le recours ultime au dictionnaire quand on ignore le sens d'un mot. A part pour ce dernier point, l'acquisition met au premier plan la notion de contexte, ce qui semble également conforme à la réalité observée. On remarquera juste que la notion de contexte ne se limite pas au corpus dans la réalité, d'où le recours nécessaire à une interactivité plus forte quand on procède à une acquisition simplement à base de corpus, par rapport au processus d'acquisition du langage chez l'enfant, d'une grande complexité et où l'information multimodale joue un rôle primordial.

**Bibliographie**